\documentclass[conference]{IEEEtran}
\IEEEoverridecommandlockouts
\usepackage{multirow}
\usepackage{cite}
\usepackage{booktabs}
\usepackage{amsmath,amssymb,amsfonts}
\usepackage{algorithmic}
\usepackage{graphicx}
\usepackage{lipsum}
\usepackage{textcomp}
\usepackage{xcolor}
\usepackage{comment}
\usepackage{hyperref}
\usepackage{subcaption}
 \hypersetup{
colorlinks=true,
linkcolor=blue,
filecolor=blue,
citecolor = blue,
urlcolor=blue,
}
\def\BibTeX{{\rm B\kern-.05em{\sc i\kern-.025em b}\kern-.08em
    T\kern-.1667em\lower.7ex\hbox{E}\kern-.125emX}}
\begin{document}
\title{Enhancing Multi-Class Disease Classification: Neoplasms, Cardiovascular, Nervous System, and Digestive Disorders Using Advanced LLMs\\
}

\author{
\IEEEauthorblockN{Ahmed Akib Jawad Karim}
\IEEEauthorblockA{\textit{Department of Computer Science and Engineering} \\
\textit{BRAC University}\\
Dhaka, Bangladesh \\
akibjawaad@gmail.com}
\and
\IEEEauthorblockN{Muhammad Zawad Mahmud}
\IEEEauthorblockA{\textit{Department of Electrical and Computer Engineering} \\
\textit{North South University}\\
Dhaka-1229, Bangladesh \\
zawad.mahmud1@northsouth.edu}
\and
\IEEEauthorblockN{Samiha Islam}
\IEEEauthorblockA{\textit{Department of Electrical and Computer Engineering} \\
\textit{North South University}\\
Dhaka-1229, Bangladesh \\
samiha.islam2@northsouth.edu}\\
\and
\IEEEauthorblockN{Aznur Azam}
\IEEEauthorblockA{\textit{Department of Computer and Science Engineering} \\
\textit{Bangladesh Army University of Science and Technology}\\
Saidpur, Bangladesh \\
aznurazam2@gmail.com
}

}


\maketitle

\begin{abstract}
In this research, we explored the improvement in terms of multi-class disease classification via pre-trained language models over Medical-Abstracts-TC-Corpus that spans five medical conditions. We excluded non-cancer conditions and examined four specific diseases. We assessed four LLMs, BioBERT, XLNet, and BERT, as well as a novel base model (Last-BERT). BioBERT, which was pre-trained on medical data, demonstrated superior performance in medical text classification (97\% accuracy). Surprisingly, XLNet followed closely (96\% accuracy), demonstrating its generalizability across domains even though it was not pre-trained on medical data. LastBERT, a custom model based on the lighter version of BERT, also proved competitive with 87.10\% accuracy (just under BERT's 89.33\%). Our findings confirm the importance of specialized models such as BioBERT and also support impressions around more general solutions like XLNet and well-tuned transformer architectures with fewer parameters (in this case, LastBERT) in medical domain tasks.
\end{abstract}


\section{Introduction}
The widespread of information and the internet has led to a huge growth in the content volume of electronic documents posted on the internet. Such large and free-form text is easy to use for automatic text classification~\cite{zhou2015c}. The most common approach is called a bag of words, and binary (on a scale of 0 or 1) are features that can then be utilized in supervised classification algorithms such as Support Vector Machines (SVMs), Naive Bayesian Classifiers (Turbo \& Taxman / NHLBI ), etc.…~\cite{joachims1998text}. Given the relative sparsity and simplicity with which some phrases can be dismissed, as well as little training data, research has turned toward focusing on more complex traits. The text classification, especially the medical test classification in the field of text mining, gets more attention, as it serves with a large dataset on medical records and literature~\cite{aronson2001effective}. 
One of the most important NLP areas is text classification, which helps in assigning a set of documents to the correct categories based on their content. The latest developments in the domain of NLP (Natural Language Processing) have completely changed how text categorization is implemented as a part of medical activities. Thanks to word embeddings, transformers, and deep learning architectures (i.e., the backbone of research in NLP), systems can now categorize medical texts more accurately and with higher efficiency than ever before! Word embeddings like Word2Vec and GloVe can elucidate medical lexicon learning by revealing semantic relations among words. More recently, transformer-based architectures (most notably BERT = Bidirectional Encoder Representations from Transformers) demonstrated an extraordinary ability to deal with the intricacies of natural language as well as capturing context~\cite{devlin2018bert}. These advanced natural language processing methods used to alleviate the issue of ambiguity and polysemy in medical text seem promising. For instance, transformer-based models do better than we are at solving the select few about-counts by letting their whole decision-making process consider the surrounding context to differentiate between noise and names of different medical problems that sound exactly like or similar to symptoms. These models are also able to find intricate patterns from big datasets, hence the detection of new or uncommon medical conditions.

The aim of the research is to see how state-of-the-art natural language processing systems perform in identifying diseases from text data. Furthermore, we will compare multiple advanced NLP models and discuss which of their capabilities are more suitable for medical text classification. We will also consider the implications for public health surveillance and patient care in healthcare settings. Thus, the original contributions of this work are as follows:
\begin{itemize}
    \item For this dataset, our model BioBERT blasted all the available models and ended up with an accuracy rate of 97\%. Also, based on Table~\ref{tab:RC}, this model outperformed existing text classification models for medical conditions.
    \item Our novelty is annotated with the introduction of our custom language model, LastBERT which we trained as a smaller version of BERT. The model achieved an accuracy of 87.06 \%, which is very comparable to the BERT performance at 89.32\% but with only 29M parameters vs the 110M in the case of BERT. This model of transformers used fewer computational resources producing satisfactory results.
\end{itemize}

\section{Related Work}
In recent years, NLP has seen impressive advances with the development of powerful (pre-trained) language models like BERT and all its extensions (SciBERT, etc.), such as BERT. These models achieved remarkable performance in different tasks that concerned text classification, such as medical diagnosis or disease classification. Prior studies have demonstrated the utility of these models in automatically recognizing and categorizing diseases, suggesting that they could strengthen multi-class disease classification for major categories (e.g., neoplasms), as well as some other types such as cardiovascular, nervous system, or digestive diseases. Blom \cite{blom2023building} in her MS thesis building a conversational agent with rasa to enrich a medical abstracts dataset used the same dataset for the text classification part. She used one ML and two LLMs and got the highest accuracy from SciBERT, which is 65\%. Prabhakar and Won \cite{prabhakar2021medical} classified medical text using hybrid deep learning models. The Hallmarks dataset and AIM dataset were utilized in this study. The datasets are a collection of biomedical paper abstracts with cancer hallmark annotations and biomedical publication abstracts, which were written around 1852. This set of data includes three characteristics of cancer, including initiating metastasis and invasion, cellular energetics, and inflammation that is promoted by tumors. Their hybrid BiGRU performed remarkably with 95.76\%. 
Ahmed et al. \cite{ahmed2020classification} used deep neural networks to classify biomedical texts for cardiovascular diseases. They used the OHSUMED-400 dataset, which contains abstracts of PubMed documents from 23 cardiovascular disease classes. This model, ultimately a DNN with BLSTM, achieved 49.4\% accuracy on their test set of spectrograms. 
Chaib et al. \cite{chaib2022gl} did multi-label text classification of cardiovascular disease reports based on the GL-LSTM model. For this study, they utilized a public dataset called Ohsumed. The master file of the Ohsumed database includes 50,216 medical summaries for patients in the year 1991 and consists of nearly half a million words from unstructured text. The collection is partitioned into three subsets--training (10 percent), testing (51 out of every hundred documents are used), and background distribution (39/100 records). The model already reached 92.7\% accuracy.
Cui et al. \cite{cui2024multi} studied multi-label text classification of cardiovascular drug attributes based on BERT and BiGRU. The data was collected from NMPA. After analysis of ablation and crossover experiments, their proposed model achieved an accuracy of 83.39\%. Hagan et al. \cite{hagan2021comparison} classified cardiovascular disease using ML methods. They used two publicly available datasets. The first one is the arrhythmia dataset from the repository of machine learning databases provided by the University of California Irvine (UCI), and the second one is the Kaggle cardiovascular disease dataset. For the UCI dataset, ExtraTrees outperformed all with 96\% accuracy, with gradient boosting achieving 94\%. Although, for the Kaggle dataset, gradient boosting was able to come on top with 74\% accuracy. Kanwa et al. \cite{kanwal2023optimized} classified cardiovascular disease using ML methods. They used the cardiovascular disease dataset from Kaggle. The dataset contains a total of 70,001 records. Among the ML models, the XGBOOST and Naïve  Bayes performed well with 92.34\% and 92.31\% accuracy, respectively. 



\section{Methodology}
\subsection{Dataset}
The Medical-Abstracts-TC-Corpus~\cite{schopf2022evaluating} was used in this investigation available at GitHub. This is a text dataset on five various medical conditions. The five conditions are neoplasms, digestive system diseases, nervous system diseases, cardiovascular diseases, and general pathological conditions. These are labeled from 1 to 5 on the above sequence, respectively. There were 14,438 records, among which 3,163 were neoplasms, 1,494 were digestive system diseases, 1,925 were nervous system diseases, 3,051 were cardiovascular diseases, and 4,805 were general pathological conditions. As the general pathological condition is not infected, we decide to exclude it. The data was split into 80-20 ratios for training and testing, respectively. A total of 11,550 records were taken for training and 2,888 for testing. Table~\ref{tab:data_summary} represents a partial medical abstract of each medical condition of the dataset. 

\begin{table}[htbp]
\caption{Dataset Representation (Summary)}
\centering
\begin{tabular}{|p{0.07\linewidth}|p{0.17\linewidth}|p{0.6\linewidth}|}
\hline
\textbf{Label} & \textbf{Disease} & \textbf{Medical Abstract (Truncated)} \\
\hline
1 & Neoplasms & Neuropeptide Y and neuron-specific enolase levels in benign and malignant pheochromocytomas. Neuron-specific enolase (NSE) is the isoform of enolase, a glycolytic enzyme found in the neuroendocrine system... \\
\hline
2 & Digestive system diseases & Sexually transmitted diseases of the colon, rectum, and anus. The challenge of the nineties. During the past two decades, an explosive growth in both the prevalence and types of sexually transmitted diseases... \\
\hline
3 & Nervous system diseases & Does carotid restenosis predict an increased risk of late symptoms, stroke, or death? The identification of carotid restenosis as an unexpected late complication of carotid endarterectomy has prompted concerns regarding its importance as a source of new cerebral symptoms, stroke, and death.... \\
\hline
4 & Cardiovascular diseases & Pharmacomechanical thrombolysis and angioplasty in the management of clotted hemodialysis grafts: early and late clinical results. The results of pharmacomechanical thrombolysis and angioplasty of 121 thrombosed hemodialysis grafts... \\
\hline
\end{tabular}
\label{tab:data_summary}
\end{table}

\subsection{Data Prepossessing}
As the database was imbalanced, we used the up-sampling and down-sampling techniques to balance it. Among the training set, data was divided into 80:20 ratio again. Here, 80\% was used for training, and 20\% validation set was used for validation. The test set was used to evaluate the models after training was completed. Finally, pandas DataFrames were converted into Hugging Face datasets for training, validation, and testing.
\subsection{Training Methodology}
\subsubsection{BioBERT}
Hugging Face model BioBERT (biobert-base-cased-v1.1) was loaded along with the BioBERT tokenizer and model for sequence classification, as well as a function to tokenize medical abstracts applying padding/truncation so they all have the same size. These functions then tokenize the datasets in a batched fashion for training or evaluation. Training arguments were epoch = 10, batch size = 16, warmup steps = 100, early stopping patience = 3, and weight decay of 0.1, respectively, for training.
\subsubsection{XLNet}
The XLNetTokenizer and model were configured for sequence classification, i.e., the 'xlnet-base-cased' type. A function to tokenize medical abstracts, pad, and truncate it as a maximum length of 512 tokens. The image processing function is then applied in a batched manner to the training, validation, and test datasets to get them ready based on what comes next: model training evaluation. Setting the training arguments as epoch = 10, early stopping patience = 2, weight decay=  0.1, and batch size=16 with warm-up steps =100 for training of pre-train BioBERT model.
\subsubsection{BERT}
We have loaded a pre-trained BERT Tokenizer (for 'bert-base-cased' model). An anonymization function is then written to prepare our medical abstracts for the tokenizer by trimming and padding them, readying each input length of 512 tokens. This tokenization function is then applied to the train, validation, and test datasets in a batched manner to preprocess them for model training and evaluation. The training arguments are epoch = 10, batch size = 16, warmup steps = 100, early stopping patience = 2, and weight decay 0.1 for training the model.
\subsubsection{LastBERT}
BERT tokenizer was created, and the model 'bert-base-uncased' was loaded. It has a tokenization function, which is also specific to the task of processing medical abstracts, padding and truncating each "abstract" -- which guarantees that all are 512 tokens long. The $CustomBERTModel$ class is kind of a wrapper over the base BERT model (we use BertForSequenceClassification), and we simply add a dropout layer on top of Bert's output so as to avoid Overfitting. The forward method implements both the calculation of loss and logits in it. This initializes the custom model as a function $createcustomstudentmodel$ that tweaks configuration for smaller BERT models by tuning parameters such as hidden size, num attention heads, num of hidden layers, and intermediate size. After that, this customized model will be transferred to the available P100 GPU for training/inference. The training was done with the following training arguments: number of epochs=10, batch size=16, warm-up steps = 100, weight decay is decaying rate = 0.1, and early stopping patience =2 \cite{karim2024largermodelsyieldbetter}.
\subsection{Work Flow Diagram}
\begin{figure}[htbp]
    \centering
    \includegraphics[width=0.5\textwidth]{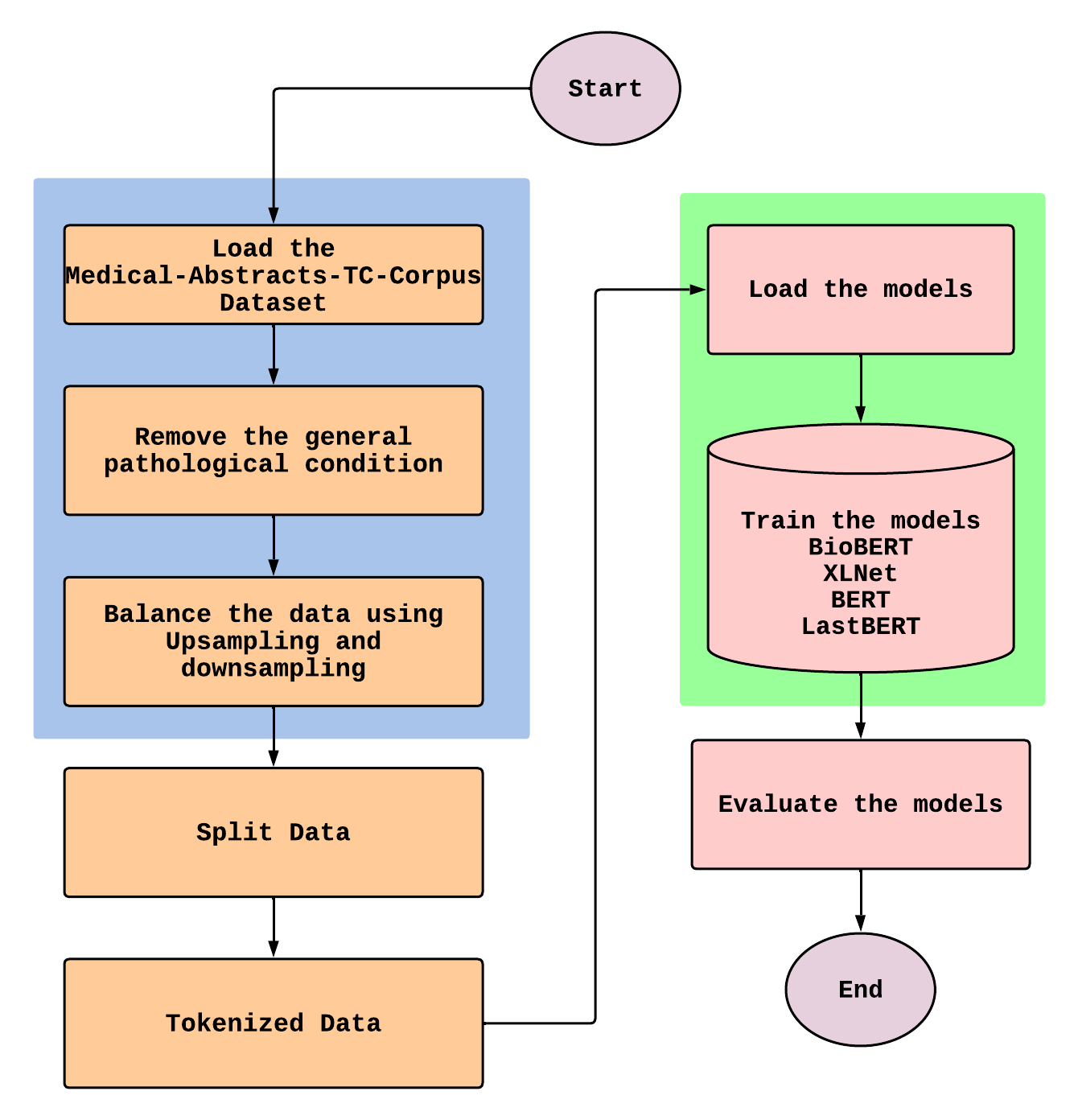}
    \caption{Workflow diagram of the proposed study of multi-class disease classification}
    \label{fig:wf}
\end{figure}

The workflow diagram of this study is visualized in Fig.~\ref{fig:wf}

\section{Result Analysis}
In this section, model training curves, performance curves, confusion matrix, and ROC curve for the two best models with the worst model's training curves and confusion matrix are shown.

\subsection{BioBERT}
\begin{figure}[htbp]
    \centering
    \includegraphics[width=0.4\textwidth]{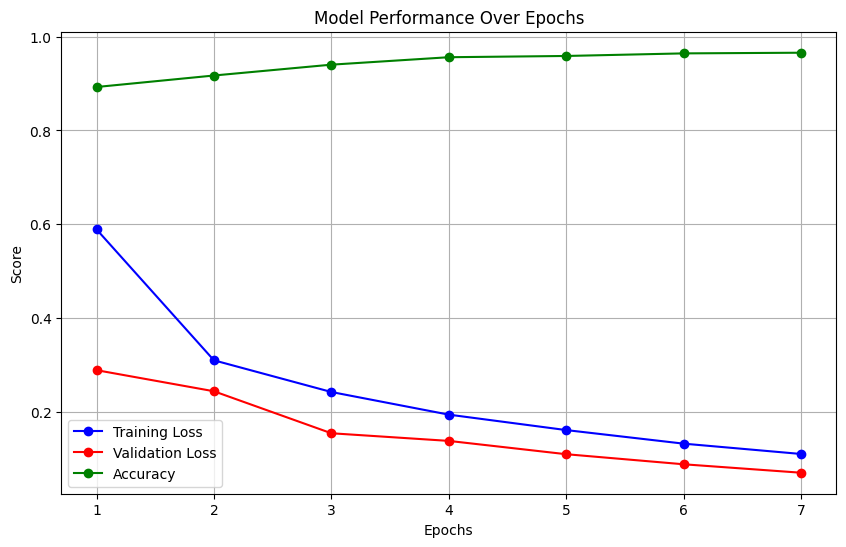}
    \caption{Loss and accuracy curve of BioBERT}
    \label{fig:LO_BBE}
\end{figure}

Fig.~\ref{fig:LO_BBE} displays the training loss and accuracy curve for the BioBERT Model. The training and validation losses decrease steadily, which is a parameter for effective learning as well as improved generalization. Meanwhile, accuracy rises to about 1.0 and remains there confirming the performance of being able to classify those medical conditions becomes extremely stable as training goes on. 

\begin{figure}[htbp]
    \centering
    \includegraphics[width=0.4\textwidth]{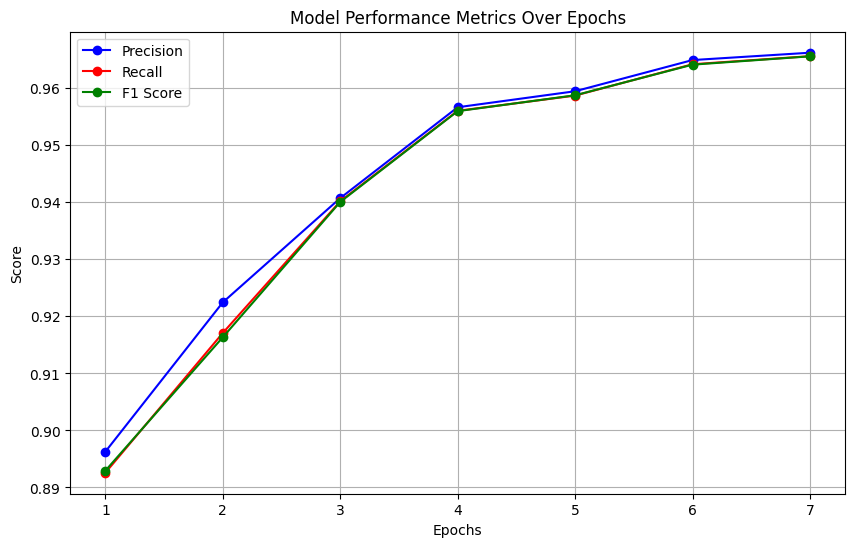}
    \caption{Performance curve of BioBERT}
    \label{fig:PG_BBE}
\end{figure}

Fig.~\ref{fig:PG_BBE} represents the performance curve of BioBERT model. The three metrics demonstrate an acceptable smooth behavior over the epochs, from around 0.89–0.90 in the first epoch to roughly 0.96 by (7/12). This means that performance improves, without loss in accuracy or reliability as the model is trained further into more epochs.

\begin{figure}[htbp]
    \centering
    \includegraphics[width=0.4\textwidth]{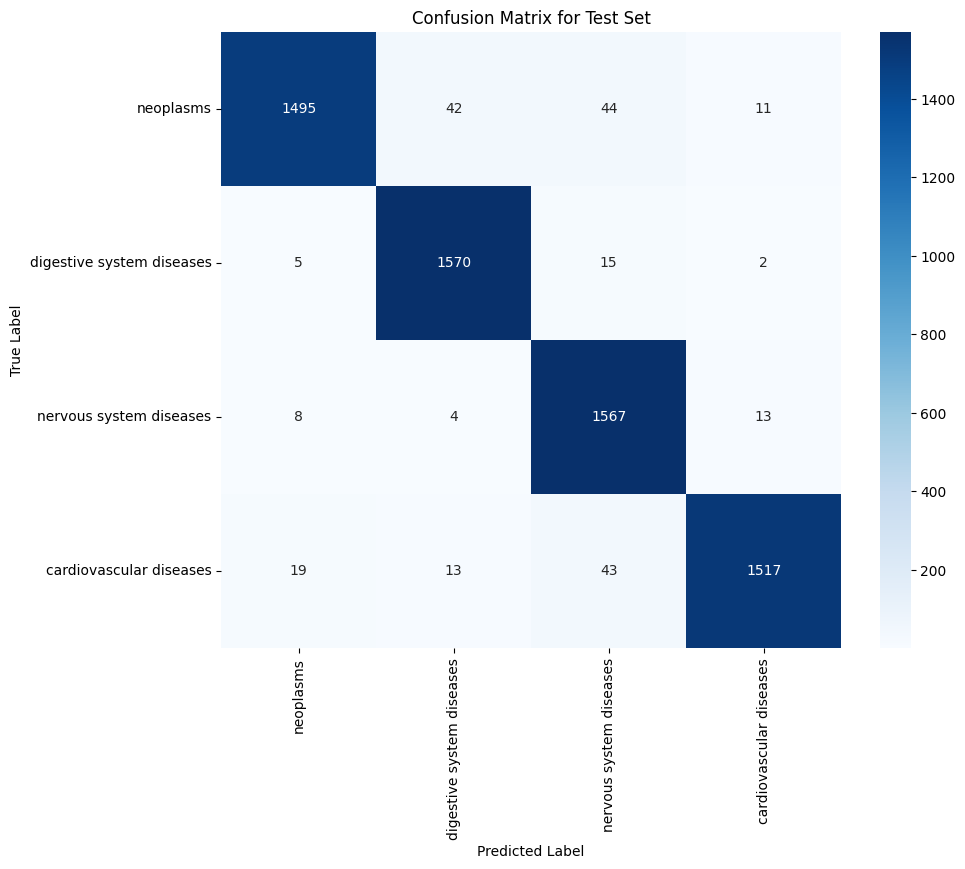}
    \caption{Confusion matrix of BioBERT}
    \label{fig:CM_BBE}
\end{figure}

In Fig. \ref{fig:CM_BBE}, the confusion matrix of the BioBERT is displayed. The matrix is quite accurate with a large number of correct predictions being along the diagonal which means that the model was able to correctly classify many samples in each class. A few cases of neoplasms were wrongly labeled as other diseases, thus the generalization power and robustness for a text classification task are excellent. 

\begin{figure}[htbp]
    \centering
    \includegraphics[width=0.4\textwidth]{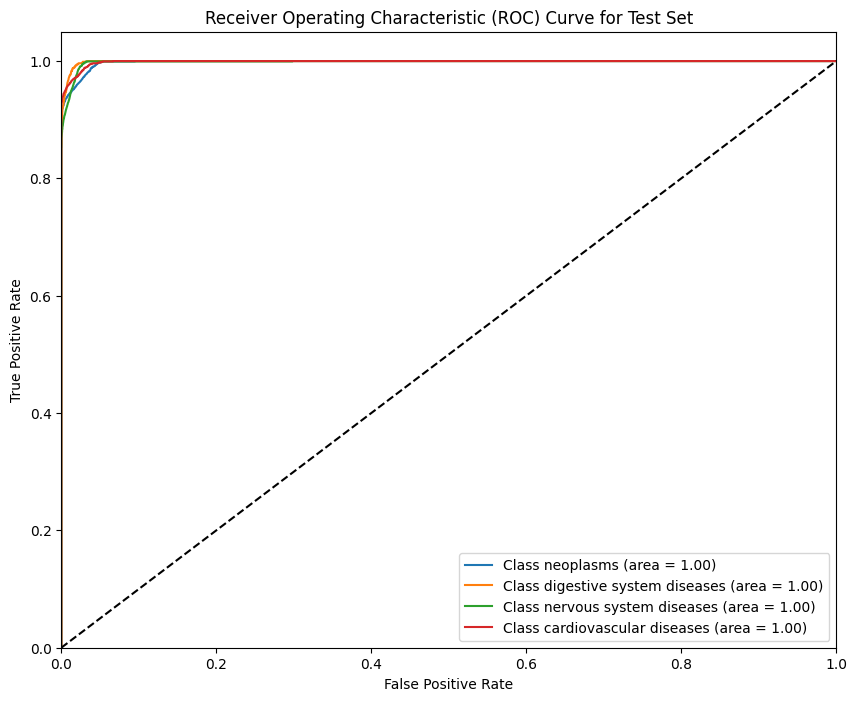}
    \caption{ROC curve of BioBERT}
    \label{fig:ROC_BBE}
\end{figure}

The ROC curve for the BioBERT model is visualized in Fig.~\ref{fig:ROC_BBE}. All classes get an AUC of 1.00 because few false positives appear in the classification performance. The curve displays the capability of that model to differentiate between various healthcare classes in a dataset.

\subsection{XLNet}

\begin{figure}[htbp]
    \centering
    \includegraphics[width=0.4\textwidth]{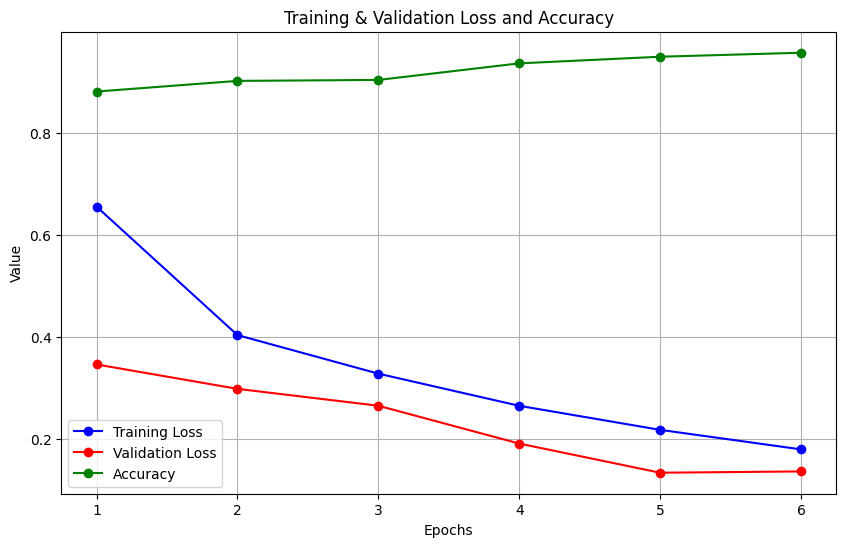}
    \caption{Loss and accuracy curve of XLNet}
    \label{fig:LO_XL}
\end{figure}

In Fig.~\ref{fig:LO_XL}, the training loss and accuracy graph for the XLNet model is shown. The training and validation losses are always heading down which is also a good point that the model learned something new so previously made errors decreased. At the same time, accuracy also continuously grows to over 0.9, which means better ability in each epoch to detect medical conditions correctly.

\begin{figure}[htbp]
    \centering
    \includegraphics[width=0.4\textwidth]{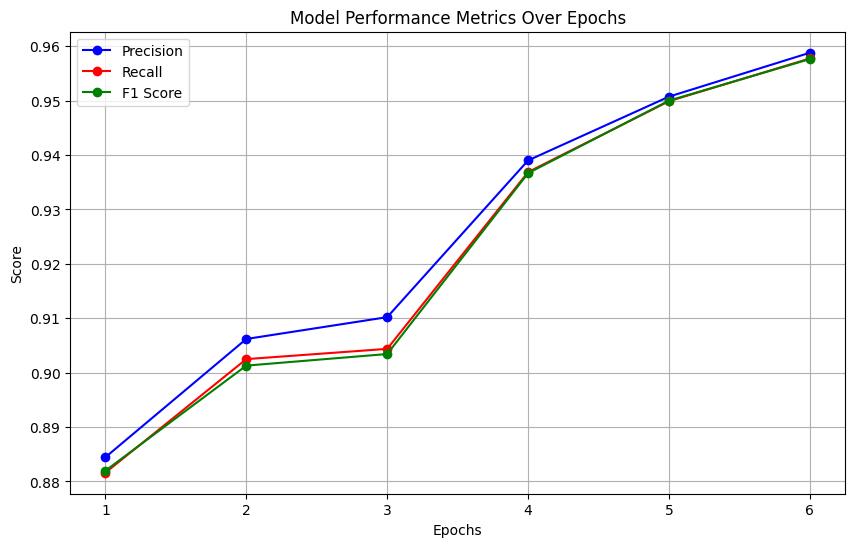}
    \caption{Performance curve of XLNet}
    \label{fig:PG_XL}
\end{figure}

Fig.~\ref{fig:PG_XL} displays the performance metrics graph for the XLNet model. The lowest score for all was 0.88-0.89 on the first epoch. The three metrics show a smooth improvement, with values nearly reaching 0.96 by epoch six while still increasing in value. This means the model is getting better and improving for text classification with every epoch by its ability to precisely categorize output classes, which in turn conclusively increases precision-recall and overall performance as it goes on training. 

\begin{figure}[htbp]
    \centering
    \includegraphics[width=0.4\textwidth]{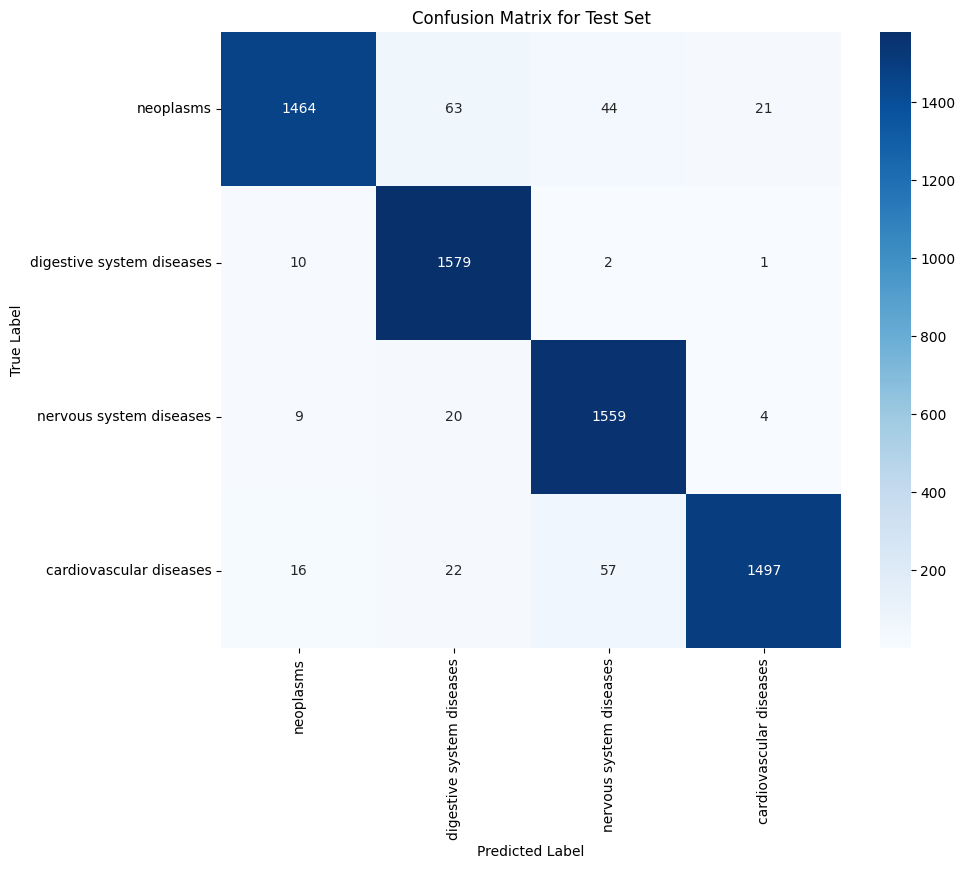}
    \caption{Confusion matrix of XLNet}
    \label{fig:CM_XL}
\end{figure}

Fig.~\ref{fig:CM_XL}, the confusion matrix of the XLNet model is shown. The matrix is highly accurate, you can see most of the predictions are along a diagonal true positive and false negative. Still, certain misclassifications do occur; a number of neoplasms were classified as other diseases, though sparsely. To sum up, the matrix shows that the XLNet model performs well in text classification tasks and is able to predict correctly for over 95\% of samples across all classes. 

\begin{figure}[htbp]
    \centering
    \includegraphics[width=0.4\textwidth]{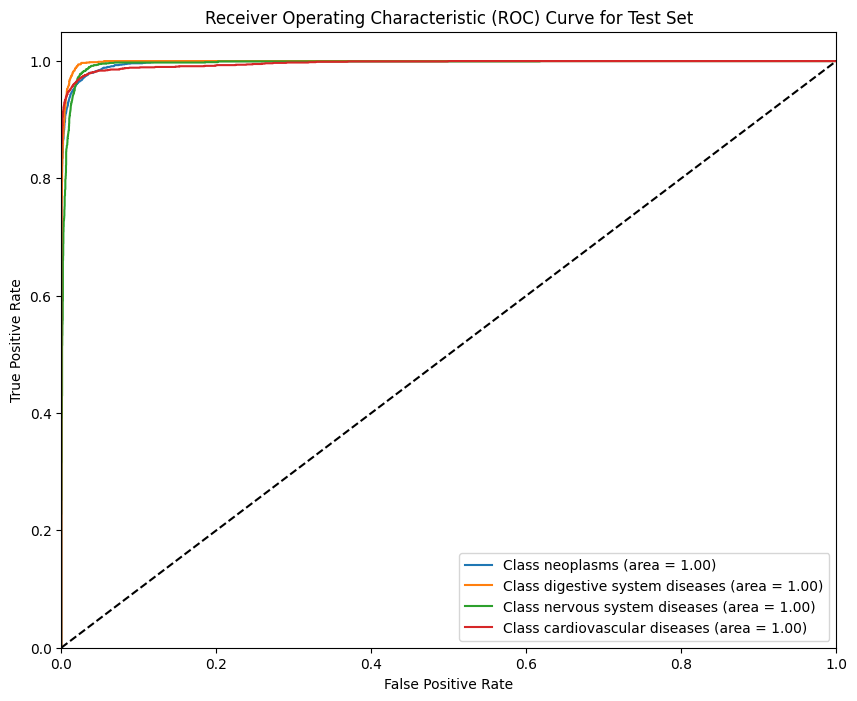}
    \caption{ROC curve of XLNet}
    \label{fig:ROC_XL}
\end{figure}

ROC curves for the XLNet model are visualized in Fig. \ref{fig:ROC_XL}. These class-averaged precision-recall curves are perfect, in the sense of having AUCs equal to 1.00, which essentially means that there is a full separation and hence few false negatives or positives at all for ideal classifiers. This graph highlights that the XLNet model performs among the best in accurately classifying different medical classes present within our dataset.

\subsection{LastBERT}
\begin{figure}[htbp]
    \centering
    \includegraphics[width=0.4\textwidth]{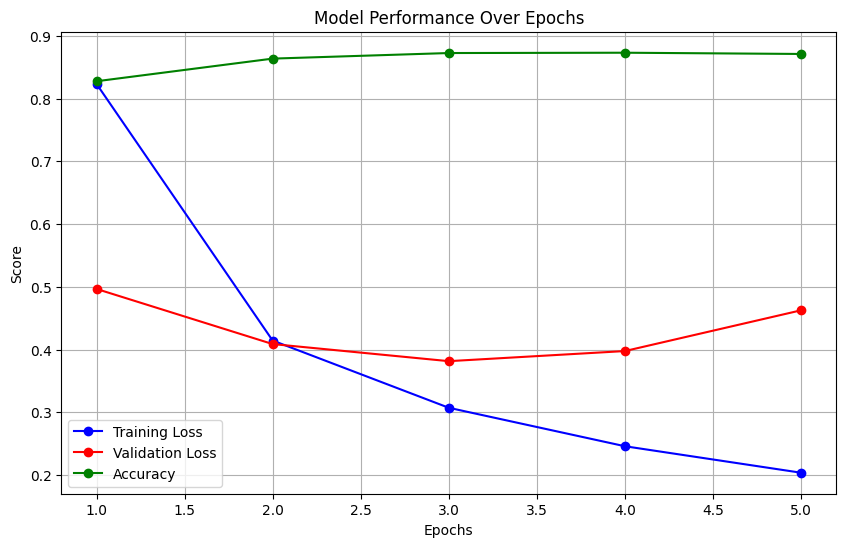}
    \caption{Loss and accuracy curve of LastBERT}
    \label{fig:LO_LBE}
\end{figure}

Fig.~\ref{fig:LO_LBE} shows the loss and accuracy curve for LastBERT. The loss in the training is steadily decreasing, demonstrating that the model keeps learning. Although validation loss decreases at the beginning, after the third epoch it starts to rise a little which could mean overfitting is in implementation. However, the accuracy is pretty constant and remains close to 0.9, so this probably means that our model performs classification very well throughout all training periods. 

\begin{figure}[htbp]
    \centering
    \includegraphics[width=0.4\textwidth]{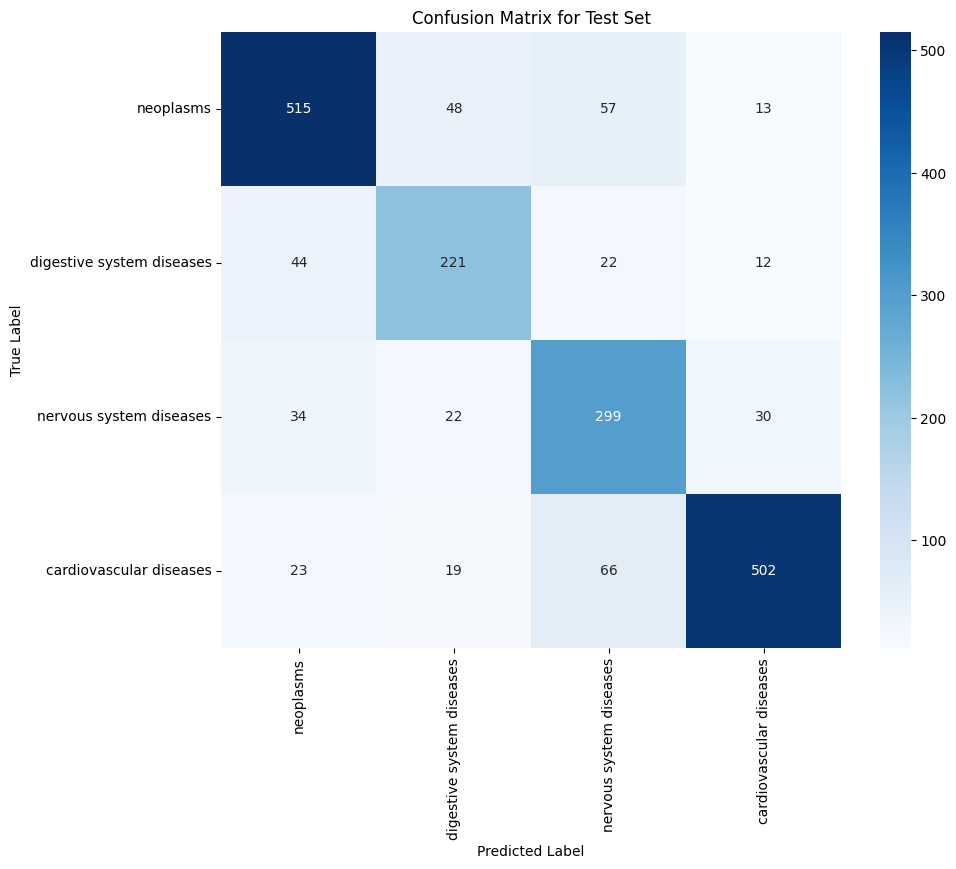}
    \caption{Confusion matrix of LastBERT}
    \label{fig:CM_LBE}
\end{figure}

The confusion matrix of the LastBERT is shown in Fig. ~\ref{fig:CM_LBE}. There is a higher concentration of true positive predictions in the diagonal on the matrix — and neoplasms, as well as cardiovascular heart disease are particularly diagnosed more correctly than often wrong ones. Nonetheless, some errors are worth noting — e.g., category confusion with nervous system diseases. The matrix shows that LastBERT works fine, but it could be improved especially in separating similar disease categories.

\subsection{Model Performance Comparison}
Table \ref{tab:performance_metrics} compares the performance metrics with our four language models (BioBERT, XLNet, BERT, and LastBERT) focusing on both macro-average columns, then weighted average over precision/recall/F1 score. BioBERT and XLNet show a good performance, which has more than 0.96 in all metrics (positive means classifying). Compared to these baseline performance levels, BERT and LastBERT have worse metrics overall, especially the latter, which has been shown only as good for lower classification accuracy numbers.

\begin{table}[htbp]
\scriptsize
\caption{Performance Metrics Comparison}
\centering
\begin{tabular}{|l|c|c|c|c|c|c|}
\hline
\multirow{2}{*}{\textbf{Models}} & \multicolumn{3}{c|}{\textbf{Macro Average}} & \multicolumn{3}{c|}{\textbf{Weighted Average}} \\
\cline{2-7}
 & \textbf{Precision} & \textbf{Recall} & \textbf{F1 Score} & \textbf{Precision} & \textbf{Recall} & \textbf{F1 Score} \\
\hline
BioBERT & 0.97 & 0.97 & 0.97 & 0.97 & 0.97 & 0.97 \\
\hline
XLNet   & 0.96 & 0.96 & 0.96 & 0.96 & 0.96 & 0.96 \\
\hline
BERT    & 0.81 & 0.82 & 0.81 & 0.83 & 0.82 & 0.83 \\
\hline
LastBERT & 0.78 & 0.79 & 0.78 & 0.81 & 0.80 & 0.80 \\
\hline
\end{tabular}
\label{tab:performance_metrics}
\end{table}

Table~\ref{tab:model_results} shows the number of parameters of the models in millions, accuracy, F1 score, precision, and recall of the four applied LLMs in this study. Clearly, in terms of accuracy and f1 score, BioBERT outperformed all the other models. Although XLNET was not trained on biomedical datasets, it still performed well. Although it is the smaller model, lastBERT produced respectable results with only 29M compared to all the others. This means faster training in a low computational resource. BERT is a 110M large LLM.

\begin{table}[htbp]
 \scriptsize
\caption{Model's Results}
\centering
\begin{tabular}{|c|c|c|c|c|c|}
\hline
\textbf{Models} & \textbf{Parameters} & \textbf{Accuracy} (\%) & \textbf{F1 Score} & \textbf{Precision} & \textbf{Recall}\\
\hline
BioBERT & 110M & 97.00 & 0.9656 & 0.9662 & 0.9656\\
\hline
XLNet & 110M & 96.00 & 0.9577 & 0.9588 & 0.9578\\
\hline
BERT & 110M & 89.33 & 0.8932 & 0.8943 & 0.8933\\
\hline
LastBERT & 29M & 87.10 & 0.8706 & 0.8943 & 0.8722\\
\hline
\end{tabular}
\label{tab:model_results}
\end{table}

As shown in Table~\ref{tab:RC}, the models are compared to those previously studied. It is evident from the table that the BioBERT model overpowers all others in the framework.

\begin{table}[htbp]
\caption{Result Comparison}
\centering
\begin{tabular}{|c|c|c|c|c|}
\hline
\textbf{Study} & \textbf{Dataset} & \textbf{Model} & \textbf{Accuracy (\%)} \\
\hline
This paper & Medical-Abstracts- & BioBERT & \textbf{97.00} \\
           & TC-Corpus & & \\
\hline
This paper & Medical-Abstracts- & XLNet & \textbf{96.00} \\
           & TC-Corpus & & \\
\hline
\cite{hagan2021comparison} & Arrhythmia dataset & ExtraTrees & 96.00 \\
        & from UCI & & \\
\hline
\cite{prabhakar2021medical} & Hallmarks dataset & BiGRU & 95.76 \\
 & and AIM dataset & & \\
\hline
\cite{chaib2022gl} & Ohsumed & GL-LSTM & 92.70 \\
\hline
\cite{kanwal2023optimized} & Cardiovascular Disease & XGBoost & 92.34 \\
 & dataset from Kaggle & & \\
\hline
\end{tabular}
\label{tab:RC}
\end{table}
\section{Conclusion and Future Work}
Advanced large language models are used in this study to classify text abstracts to determine which medical condition the person is in. It is not surprising that BioBERT pre-trained with medical text performed best (97\% accuracy) among the models, demonstrating once again how essential domain-specific pre-trained model training is to medical-text classification tasks. An extremely strong performance of 96\% fresh normal precision, suggesting that general-purpose models still may provide decent quality in the at-spontaneous field when finetuned (XLNet not even created for the medical domain; yet?) Although the custom-developed LastBERT model was smaller and more resource-efficient, It achieved a comparable performance against a much larger baseline BERT, which suggests that even with content resources, the use of purpose-optimized models could help. Several future works can be performed to enhance the accuracy and efficiency of the model, and the following potential options could be considered for experimentation. One possible first step in this direction would be to experiment with including more domain-specific data at training time from which models like XLNet and LastBERT could benefit. This could potentially lead us down a second path of avant-garde algorithms, or maybe the answer is to do more work on hybrid, so as you get many benefits from one methodology and others benefitting in another, then at least have this slant going for classification can be pretty solid. Another benefit of this research is the fact that it reduces model bloat, making them actually usable, which could come in handy for deploying solutions more extensively, including resource-limited healthcare settings.

\bibliographystyle{IEEEtran}
\bibliography{IEEE.bib}

\end{document}